\documentclass[10pt,twocolumn,letterpaper]{article}

\usepackage[pagenumbers]{wacv} 
\usepackage{graphicx}
\usepackage{amsmath}
\usepackage{amssymb}
\usepackage{booktabs}
\usepackage{glossaries}
\usepackage{multirow}
\usepackage{tikz}
\usepackage{pgfplots}
\usepackage{KAcolors}
\usepackage[accsupp]{axessibility}

\pgfplotsset{
	compat=1.14,%
	plot coordinates/math parser=false,%
	tick label style={font=\footnotesize},%
	label style={font=\small},%
	scale only axis,%
	axis lines=center,%
	axis on top,%
	every axis legend/.append style={cells={anchor=west},draw=none,font=\small},%
	every axis plot/.append style={semithick},%
	KIT scatter plot A/.style={%
			draw=none,only marks,mark=*,mark options={draw=KITblue,fill=KITblue}},%
	KIT scatter plot B/.style={%
			draw=none,only marks,mark=square*,mark options={draw=KITred,fill=KITred}},%
	KIT scatter plot C/.style={%
			draw=none,only marks,mark=diamond*,mark options={draw=KITorange,fill=KITorange}},%
	KIT scatter plot explicit/.style={%
			scatter,%
			scatter/classes={%
					a={mark=*,KITblue},%
					b={mark=square*,KITred},%
					c={mark=diamond*,KITorange}},%
			only marks,%
			scatter src=explicit symbolic,%
			z buffer=sort},%
	KIT ybar plot A/.style={%
			ybar,fill=KITblue,draw=none},%
	KIT ybar plot B/.style={%
			ybar,fill=KITred,draw=none},%
	KIT ybar plot C/.style={%
			ybar,fill=KITorange,draw=none},%
	KIT xbar plot A/.style={%
			ybar,fill=KITblue,draw=none},%
	KIT xbar plot B/.style={%
			ybar,fill=KITred,draw=none},%
	KIT xbar plot C/.style={%
			ybar,fill=KITorange,draw=none},%
	KIT line plot A/.style={%
			KITblue,semithick},%
	KIT line plot B/.style={%
			KITred,semithick},%
	KIT line plot C/.style={%
			KITorange,semithick},%
	KIT line plot D/.style={%
			KITlilac,semithick},%
	KIT line plot E/.style={%
			KITbrown,semithick},%
	KIT line plot F/.style={%
			KITblue,semithick,dashed},%
	KIT line plot G/.style={%
			KITred,semithick,dashed},%
	KIT line plot H/.style={%
			KITorange,semithick,dashed},%
	KIT line plot I/.style={%
			KITlilac,semithick,dashed},%
	KIT line plot J/.style={%
			KITbrown,semithick,dashed},%
	KIT smooth plot A/.style={%
			KITblue,semithick,smooth},%
	KIT smooth plot B/.style={%
			KITred,semithick,smooth},%
	KIT smooth plot C/.style={%
			KITorange,semithick,smooth},%
	KIT smooth plot D/.style={%
			KITlilac,semithick,smooth},%
	KIT smooth plot E/.style={%
			KITbrown,semithick,smooth},%
	KIT smooth plot F/.style={%
			KITblue,semithick,dashed,smooth},%
	KIT smooth plot G/.style={%
			KITred,semithick,dashed,smooth},%
	KIT smooth plot H/.style={%
			KITorange,semithick,dashed,smooth},%
	KIT smooth plot I/.style={%
			KITlilac,semithick,dashed,smooth},%
	KIT smooth plot J/.style={%
			KITbrown,semithick,dashed,smooth},%
}

\usepackage[pagebackref,breaklinks,colorlinks]{hyperref}

\usepackage[capitalize]{cleveref}
\crefname{section}{Sec.}{Secs.}
\Crefname{section}{Section}{Sections}
\Crefname{table}{Table}{Tables}
\crefname{table}{Tab.}{Tabs.}

\def\wacvPaperID{*****} 
\def\confName{WACV}
\def\confYear{2024}

\begin{document}

\title{Security Fence Inspection at Airports Using Object Detection}

\author{Nils Friederich\textsuperscript{1} \quad Andreas Specker\textsuperscript{3,4} \quad  
Jürgen Beyerer\textsuperscript{3,2,4} \\
\textsuperscript{1}Karlsruhe Institute of Technology, Institute for Automation and Applied Informatics\\
\textsuperscript{2}Karlsruhe Institute of Technology, Institute for Anthropomatics and Robotics\\
\textsuperscript{3}Fraunhofer IOSB \quad \textsuperscript{4}Fraunhofer Center for Machine Learning \\
{\tt\small nils.friederich@kit.edu} \tt\small \{andreas.specker,juergen.beyerer\}@iosb.fraunhofer.de}

\maketitle

\newacronym{adaboost}{AdaBoost}{Adaptive Boosting}

\newacronym{ai}{AI}{Artificial Intelligence}

\newacronym{amp}{AMP}{Automatical Mixed Precision}

\newacronym[%
  shortplural={APs},%
  longplural={Average Precisions}%
] {ap}{AP}{Average Precision}

\newacronym{atss}{ATSS}{Adaptive Training Sample Selection}

\newacronym{auc}{AUC}{Area Under The Curve}

\newacronym{bn}{BN}{Batch-Normalization}

\newacronym[
    shortplural={BBoxes},%
    longplural={Boundig Boxes}%
]{bb}{BBox}{Bounding Box}

\newacronym{clahe}{CLAHE}{Contrast Limited Adaptive Histogram Equalization}

\newacronym[
    shortplural={CNNs},%
    longplural={Convolutional Neural Networks}%
]{cnn}{CNN}{Convolutional Neural Network}

\newacronym{cpn}{CPN}{Corner Proposal Network}

\newacronym{cpu}{CPU}{Central Processing Unit}

\newacronym{coco}{COCO}{Common Objects in Contexts}

\newacronym[
    shortplural={CSPNets},%
    longplural={Cross Stage Partial Networks}%
]{csp}{CSPNet}{Cross Stage Partial Network}

\newacronym{cuda}{CUDA}{Compute Unified Device Architecture}

\newacronym{cv}{CV}{Computer Vision}

\newacronym[%
  shortplural={DCNs},%
  longplural={Deformable Convolutional Networks}%
] {dcn}{DCN}{Deformable Convolutional Network}

\newacronym[%
  shortplural={DConvs},%
  longplural={Deformable Convolutions}%
]{dconv}{DConv}{Deformable Convolution}

\newacronym{detr}{DETR}{DEtetction TRansformer}

\newacronym{dl}{DL}{Deep Learning}

\newacronym[%
  shortplural={DNNs},%
  longplural={Deep Neural Networks}%
] {dnn}{DNN}{Deep Neural Network}

\newacronym{ema}{EMA}{Exponential Moving Average}

\newacronym{fast}{FAST}{Features from Accelerated Segment Test}

\newacronym{fcl}{FCL}{Fully Connected Layer}

\newacronym{fcos}{FCOS}{Fully Convolutional One-stage Object Detector}

\newacronym{flops}{FLOPs}{Floating Point Operations}

\newacronym{fn}{FN}{False Negative}

\newacronym{fp}{FP}{False Positive}

\newacronym{fp16}{FP16}{Single-Precision Format}

\newacronym{fp32}{FP32}{Half-Precision Format}

\newacronym[
    shortplural={FPGAs},%
    longplural={Field Programmable Gate Arrays}%
]{fpga}{FPGA}{Field Programmable Gate Array}

\newacronym[
    shortplural={FPNs},%
    longplural={Feature Pyramid Networks}%
]{fpn}{FPN}{Feature Pyramid Network}

\newacronym{fps}{FPS}{Frames Per Second}

\newacronym{ga}{GA}{Genetic Algorithm}

\newacronym[
    shortplural={GPUs},%
    longplural={Graphics Processing Units}%
]{gpu}{GPU}{Graphics Processing Unit}

\newacronym{hed}{HED}{Holistically-Nested Edge Detection}

\newacronym{histequ}{HistEqu}{Histogram equalization}

\newacronym{hog}{HOG}{Histogram of oriented gradients}

\newacronym{iacs}{IACS}{IoU-aware classification scores}

\newacronym{ies}{IES}{Vision and Fusion Laboratory}

\newacronym{iou}{IoU}{Intersection Over Union}

\newacronym{ir}{IR}{InfraRed}

\newacronym{kit}{KIT}{Karlsruhe Institut of Technology}

\newacronym{knn}{KNN}{K-Nearest Neighbours}

\newacronym{loocv}{LOOCV}{Leave-One-Out Cross-Validation}

\newacronym{lr}{LR}{Learning Rate}

\newacronym{luftsig}{LuftSiG}{Luftsicherheitsgesetz}

\newacronym{luftvg}{LuftVG}{Luftverkehrsgesetz}

\newacronym{luftvzo}{LuftVZO}{Luftverkehrs-Zulassungs-Ordnung}

\newacronym{map}{mAP}{mean Average Precision}

\newacronym{ml}{ML}{Machine Learning}

\newacronym{mp}{MP}{Mixed Precision}

\newacronym[%
  shortplural={MDConvs},%
  longplural={Modulated Deformable Convolutions}%
]{mdconv}{MDConv}{Modulated Deformable Convolution}

\newacronym{mmdet}{MMDet}{MMDetection}

\newacronym{nlp}{NLP}{Natural language processing}

\newacronym[
    shortplural={NNs},%
    longplural={Neural Networks}%
]{nn}{NN}{Neural Network}

\newacronym{nms}{NMS}{Non Maximum Suppression}

\newacronym{osv}{OSV}{Omnidirectional Surface Vehicle}

\newacronym{pan}{PANet}{Path Aggregation Network}

\newacronym{rcnn}{R-CNN}{Region Based Convolutional Neural Networks}

\newacronym[
    shortplural={ResNets},%
    longplural={Residual Networks}%
]{resnet}{ResNet}{Residual Network}

\newacronym[
    shortplural={RoIs},%
    longplural={Regions of Interest}%
]{roi}{RoI}{Region of Interest}

\newacronym[
    shortplural={RSeeds},%
    longplural={Random Seeds}%
]{rseed}{RSeed}{Random Seed}

\newacronym{sgd}{SGD}{Stochastic Gradient Descent}

\newacronym{sift}{SIFT}{Scale Invariant Feature Transform}

\newacronym{sota}{SOTA}{State-of-the-Art}

\newacronym{spp}{SPP}{Spatial Pyramid Pooling}

\newacronym{sppf}{SPPF}{Spatial Pyramid Pooling Fast}

\newacronym{ssd}{SSD}{Single Shot Detector}

\newacronym{surf}{SURF}{Speeded Up Robust Features}

\newacronym[
    shortplural={SVMs},%
    longplural={Support Vector Machines}%
]{svm}{SVM}{Support Vector Machine}

\newacronym{swin}{Swin}{Shifted windows}

\newacronym{tal}{TAL}{Task Aligment Learning}

\newacronym{tide}{TIDE}{\textbf{T}oolbox for \textbf{I}dentifying \textbf{O}bject Detection \textbf{E}rrors}

\newacronym{tn}{TN}{True Negative}

\newacronym{tp}{TP}{True Positive}

\newacronym[
    shortplural={TPUs},%
    longplural={Tensor Processing Units}%
]{tpu}{TPU}{Tensor Processing Unit}

\newacronym{tood}{TOOD}{Task-aligned One-stage Object Detection}

\newacronym{yolo5}{YOLOv5}{You Only Look Once v5}

\newacronym{vfnet}{VFNet}{VarifocalNet}

\newacronym[
    shortplural={VGGs},%
    longplural={Visual Geometry Groups}%
]{vgg}{VGG}{Visual Geometry Group}

\newacronym[
    shortplural={ViTs},%
    longplural={Vision Transformers}%
]{vit}{ViT}{Vision Transformer}

\newacronym{pvoc}{PASCAL VOC}{PASCAL Visual Object Classes}
\begin{abstract}
To ensure the security of airports, it is essential to protect the airside from unauthorized access.
For this purpose, security fences are commonly used, but they require regular inspection to detect damages.
However, due to the growing shortage of human specialists and the large manual effort, there is the need for automated methods.
The aim is to automatically inspect the fence for damage with the help of an autonomous robot.
In this work, we explore object detection methods to address the fence inspection task and localize various types of damages.
In addition to evaluating four \gls{sota} object detection models, we analyze the impact of several design criteria, aiming at adapting to the task-specific challenges.
This includes contrast adjustment, optimization of hyperparameters, and utilization of modern backbones.
The experimental results indicate that our optimized \gls{yolo5} model achieves the highest accuracy of the four methods with an increase of 6.9\% points in \gls{ap} compared to the baseline.
Moreover, we show the real-time capability of the model.
The trained models are published on GitHub: \href{https://github.com/N-Friederich/airport_fence_inspection}{https://github.com/N-Friederich/airport\_fence\_inspection}.
\end{abstract}
\section{Introduction}
\label{sec:intro}
In contemporary times, airplanes have assumed a crucial role in global transportation.
Ensuring the safety of passengers, cargo, and machinery is of great importance.
This requires appropriate safety mechanisms, both onboard the aircraft and within the airport infrastructure.
Protecting sensitive areas such as the airside is a major challenge for airport operators.
In Germany, for instance, there are over 540 airfields, out of which 15 are classified as international airports according to § 27d Paragraph 1 \gls{luftvg} \footnote{https://www.gesetze-im-internet.de/ (Gesetz im Netz - Federal Ministry of Justice), Date 01/09/2023}\label{gesetz_im_inet}~\cite{dfs_statistik}.
To obtain this classification, airfields must secure their sensitive areas, including the airside, against unauthorized access by adhering to § 8 \gls{luftsig}\footref{gesetz_im_inet}.
Appropriate security fences are a common practice to protect these areas~\cite{easa_cs_adr_dsn}.
These fences must be regularly checked for damage in accordance with § 8 and § 9 \gls{luftsig}\footref{gesetz_im_inet}. 
Even minor damage to the fence potentially allows animals to enter the airfield and pose a danger to themselves, people, and machinery~\cite{easa_cs_adr_dsn}.
However, the availability of skilled human personnel to perform fence inspections is becoming increasingly limited~\cite{Burstedde2022Fachkr}. 
Therefore, exploring automated methods to monitor this real-world surveillance application, such as utilizing mobile robots with cameras for detecting damages, is highly valuable. 

\begin{figure}[t]
    \centering
    \resizebox{\linewidth}{!}{
    \includegraphics[height=4.25cm]{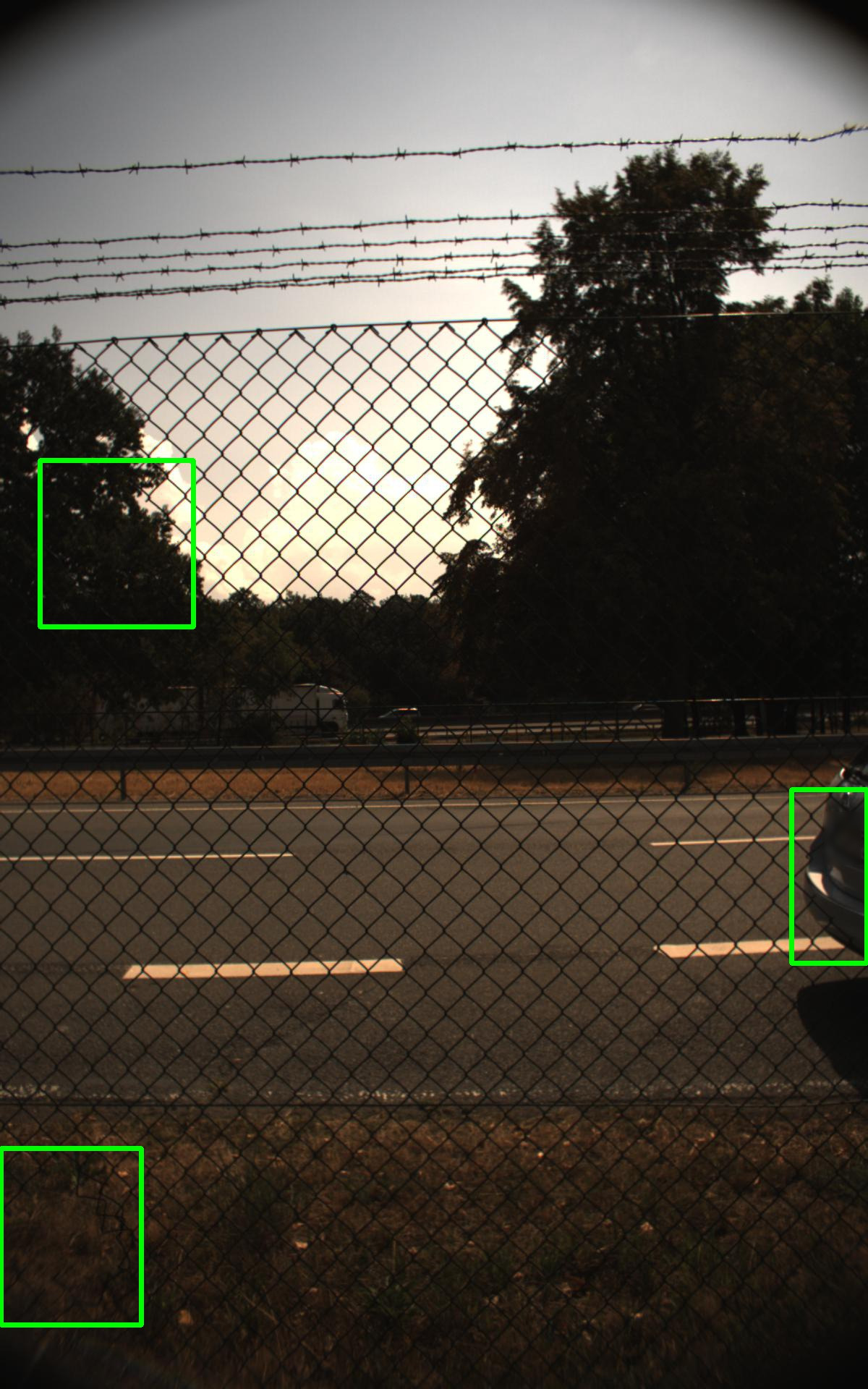}
    \hfill
    \includegraphics[height=4.25cm]{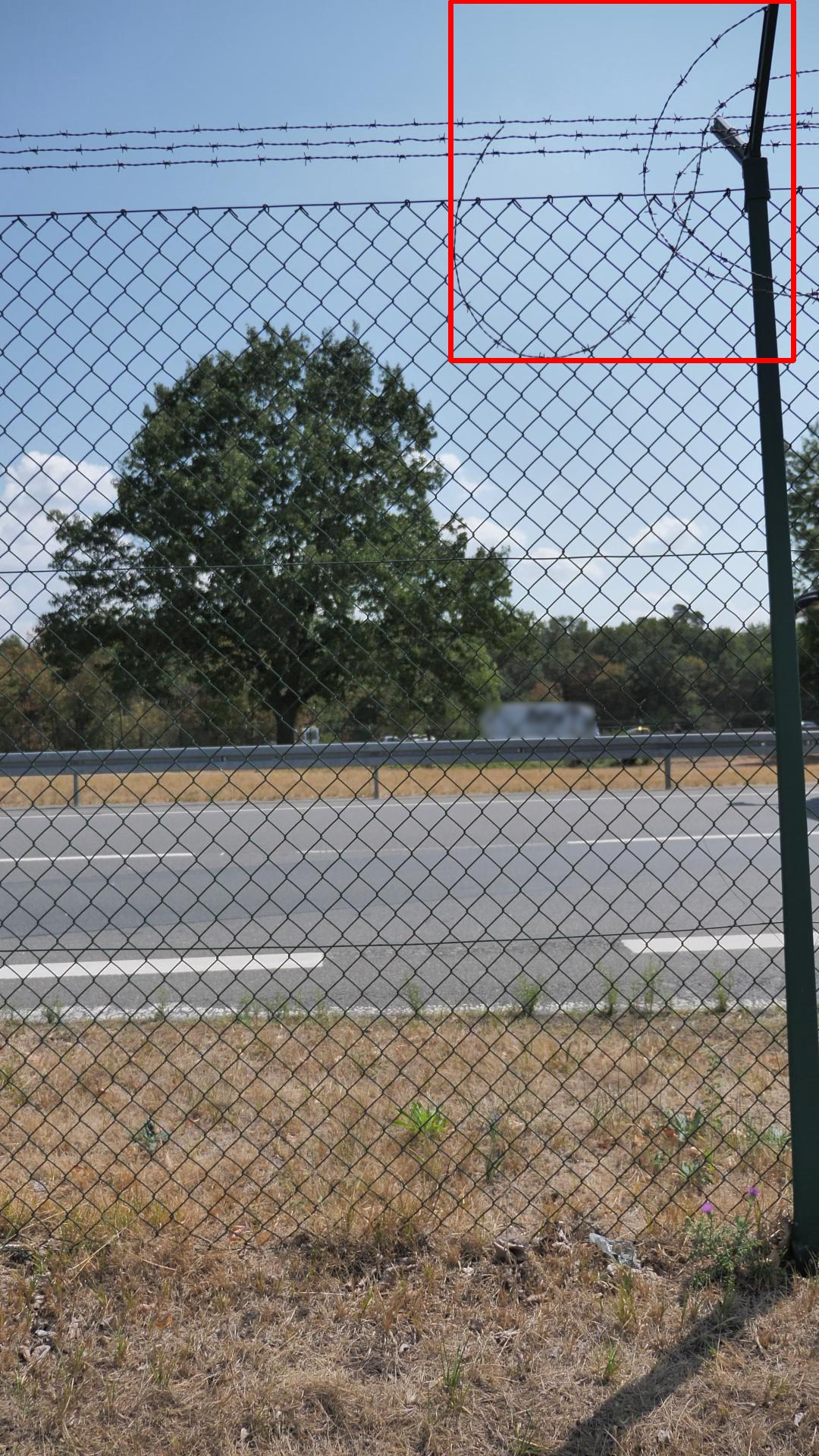}
         \label{fig:intro_example_dataset_2}
    }
    \caption{\textbf{Examples of damaged security fences} -- 
    The \gls{bb} colors symbolize different types of damage: \textcolor{green}{Green} marks a hole in the fence; \textcolor{red}{Red} marks damage to the climb-over-protection.
    }
    \label{fig:intro_example_dataset}
    \vspace{-0.3cm}
\end{figure}
To implement such an automatic system, this work focuses on 2D object detection methods for three main reasons.
First, the existing literature offers numerous robust methods to effectively tackle this task~\cite{yolov5,tood,varifocalnet,mmdetection}.
Second, using cheap camera sensors is adequate for capturing the necessary imagery.
Last, 2D image processing is computationally less heavy compared to, e.g., processing 3D data from a stereo camera.
 
In general, object detection methods aim at identifying and localizing specific objects or patterns within an input image.
In the context of this work, our objective is to detect two commonly occurring types of damages within fence images captured at airports using a self-recorded dataset. 
Two examples of airport fences are presented in~\cref{fig:intro_example_dataset}.
There is a wire mesh structure in the lower part as a passage barrier and multiple rows of barbed wire in the upper part for climbing-over protection.
Damage can occur in both sections.
However, damage detection needs a clear differentiation between the fence and structures in the background. 
Moderate contrast in many areas, such as with the trees in the background, hardens the task.
In addition, background clutter, e.g., leaves, further complicates the detection process, especially with the intricate wire mesh. 
To overcome these challenges, various techniques, including contrast adjustment, are examined throughout this work.
For this purpose, \gls{sota} deep learning methods, namely \gls{yolo5}~\cite{yolov5}, \gls{tood}~\cite{tood}, \gls{vfnet}~\cite{varifocalnet}, and Deformable \gls{detr}~\cite{deformable_detr}, are evaluated and compared for their potential in addressing the detection challenges associated with the security fence inspection task.
Ideally, the resulting detection system should work autonomously on a mobile robot.
However, this requires the most economical operation possible with reliable damage detection on affordable hardware.
Therefore, we also investigate the tradeoff between speed and accuracy. 


In summary, the main contributions of this paper are threefold:
\begin{itemize}
    \setlength{\itemsep}{-2pt}
    \item We conduct the first analysis of \gls{sota} object detection methods for the security fence inspection use case.
    \item Our thorough evaluation of various design choices highlights key factors for strong damage detection results.
    \item The resulting real-time model demonstrates remarkable performance and generalization ability and, thus, provides a strong baseline for future research.
\end{itemize}
\section{Related Work}
\label{sec:relwork}
The automated damage detection at airport fences requires \gls{cv} algorithms~\cite{cv_algos_hw_impl_survey}. 
In this use case, a simple image classification approach would be insufficient, resulting in a time-consuming search game for human operators. 
On the other hand, precise segmentation is not required for this task, as it does not demand detailed segmentation of each object instance wire. 
In addition, creating segmentation labels for intricate objects such as the wire mesh structure by human annotators would be both time-consuming and costly~\cite{pixel_label_fine_segm}. 
Therefore, object detection is utilized as a compromise between classification and segmentation.
For the purpose of object detection, \gls{dl} methods have gained prominence over classical \gls{cv} methods due to hierarchical feature extraction, higher accuracy, and improved generalization capabilities~\cite{feature_extraction_ir_cv,dl_vs_tradCV,dl_cv_brief_review,review_dl_material_degradation}. 
For object detection methods, a differentiation can be made between anchor-based and anchor-free methods. Whereas anchor-based methods often converge faster, anchor-free methods require fewer hyperparameters and may have stronger generalization capabilities. Whether this is true in the context of this thesis is evaluated using the anchor-based method YOLOv5 and the anchor-free methods \gls{tood}, \gls{vfnet} and Deformable \gls{detr}.

Regardless of the model type, \gls{dl} models often encounter issues with overfitting, particularly when dealing with small datasets.
To mitigate this issue, pre-trained models are commonly employed.
Since no pre-trained model tailored explicitly for the use case has been published, a default pre-trained model is utilized, such as those trained on the \gls{coco} dataset~\cite{coco_map,survey_tran_learn}.
Furthermore, to the best of our knowledge, no appropriate datasets for security fence inspection have been published.
Although there are related use cases, such as de-fencing~\cite{stream_img_defencing_smartphones,defencing_sing_img_dcnn,robust_effizient_defencing_cgan}, these datasets consist of images taken in closer proximity and different spatial contexts~\cite{robust_effizient_defencing_cgan}.
\section{Methodology}
\label{sec:methods}
This paper thoroughly examines the use of \gls{sota} \gls{dl} methods with different characteristics regarding their suitability for the damage detection task and derives best practices concerning design criteria.
In detail, \gls{yolo5}~\cite{yolov5}, \gls{tood}~\cite{tood}, \gls{vfnet}~\cite{varifocalnet}, and Deformable \gls{detr}~\cite{deformable_detr} are considered.
After motivating these choices in \cref{sec:methods-dl_methods}, several adaptions are introduced to increase the detection performance for the task under real-world conditions.
The overall goal is to identify the best design characteristics for \gls{dl} methods from a quantitative perspective and further investigate this method concerning the influence of input image resolution to achieve a beneficial trade-off between detection results and computational complexity.

\subsection{Deep Learning Methods}
\label{sec:methods-dl_methods}

Recently, numerous new \gls{dl} methods have been introduced~\cite{yolov5,tood,varifocalnet,ddod,detr,deformable_detr}.
In terms of real-time object detection, several derivatives of the YOLO approach~ \cite{yolov3v4v5_auto_brunch_det,  yolov3v4v5_autonom_faulty_uavs,yolov3v4v5_paultry_recognition,yolov5_face_recog} have proven suitable for various real-world applications~\cite{yolov5_safty_helmet_det,yolov5_face_mask}.
For instance, \gls{yolo5} achieves good detection results at lower operational expenses.
However, \gls{yolo5} and its predecessors~\cite{yolo9000,yolov3,yolov4} are anchor-based, which may lead to limitations in generalization capabilities~\cite{analysis_anchorbased_anchorfree_od_dl}. 
Therefore, two anchor-free \gls{dl} methods are included in the analysis, namely \gls{tood}~\cite{tood} and \gls{vfnet}~\cite{varifocalnet}.

All these three methods were developed as CNN-based methods \cite{tood,varifocalnet,yolov5}. 
Since transformer-based models promise improved generalization capabilities~\cite{cnn_vs_vit}, the transformer-based Deformable \gls{detr}~\cite{deformable_detr}, a successor of the popular \gls{vit}-based \gls{detr}~\cite{detr}, is investigated. 
However, Transformers, such as \gls{vit}, typically require more training data than \glspl{cnn}~\cite{scale_vision_transformer}. 
Since the available data for the fence inspection task is limited, further investigations need to be conducted.


\subsection{Optimizations}
\label{sec:methods-optimizations}
In this work, we thoroughly study various design parameters to improve damage detection in security fences under real-world settings.
In the following, the considered aspects are motivated and introduced.

\noindent\textbf{Numerical stability:}
When implementing \gls{dl} methods, numerical instabilities such as exploding gradients or zero divisions may occur.
These numerical instabilities can lead to a degradation of the training results, which is why we eliminate them to improve the meaningfulness of the experiments. 
We contributed our code changes to the original code repositories.

\noindent\textbf{Regularization:}
Regularization of \gls{dl} models is crucial for preventing overfitting on small datasets with few \glspl{roi} per image.
For this, primarily three adaptations are investigated. 
First, the image weighting technique from \gls{yolo5} is used to over-represent difficult training examples.
Due to the small training dataset, edge cases that occur rarely may otherwise be covered by the background noise of decent images.
Second, optimizers with regularization abilities like Adam~\cite{adam} or AdamW~\cite{adamW} are investigated. To prevent gradient oscillations but at the same time allow for a steep gradient descent, the impact of learning rate adjustments is explored.

\noindent\textbf{Data augmentation:}
Data augmentation methods aim at increasing the diversity in small-scale datasets to prevent overfitting and improve robustness.
Due to the small amount of data with few damages each, the impact of data augmentation methods like mosaic and affine transformations are investigated.

\noindent\textbf{Contrast enhancement:} 
Poor contrast, e.g., caused by low light during dusk or dawn, presents a significant challenge in detecting damages on airport fences.
In such cases, the fine structures of the fences do not stand out clearly against the background.
Pre-processing images with contrast enhancement methods prior to damage detection alleviates the problem.
Contrast adjustment can generally be executed on the entire image or separately for multiple image regions.
We compare both global and local contrast enhancement methods represented by \gls{histequ}~\cite{comprehensive_survey_img_contrast_techniques,survey_img_enhancement_tequ} and \gls{clahe}~\cite{clahe}, respectively.

\noindent\textbf{Backbone:}
While \gls{yolo5} utilizes a modern CSPDarknet~\cite{yolov4_scaled,cspnet} as backbone~\cite{yolov5}, \gls{tood} and \gls{vfnet} rely on variants of the \gls{resnet}~\cite{resnet} and ResNeXt~\cite{resnext} architectures.
However, more recent backbones such as Res2Net~\cite{res2net} or ConvNeXt~\cite{convNeXt} show better performance in various tasks~\cite{convnext_yolo,convnext_unet}. 
Therefore, these backbones are applied in conjunction with \gls{tood} and \gls{vfnet}. 
Analogous to the original backbones, we pre-train these backbones on the \gls{coco} dataset first.

\noindent\textbf{Hyperparameter tuning:}
The choice of appropriate hyperparameters is essential to assure good performance, especially if few training data are available.
In addition, the fence inspection task requires strong generalization capabilities.
Due to the different conditions and demands, hyperparameters proposed by the original works might not be optimal in damage detection.
As a result, detailed studies concerning the choice of hyperparameters are conducted.


\noindent\textbf{Image resolution:}
When object detectors are deployed in real-world applications, fast computation is crucial.
For instance, if the processing is performed on autonomous platforms, such as robots.
The inference speed of object detectors is greatly affected by the resolution of the input images.
Higher-resolution images provide a more detailed context, enabling improved detection of damages, while the computational complexity increases.
Thus, achieving a suitable trade-off between detection accuracy and computational requirements is essential.



\section{Experiments}
\label{sec:experiements}
For maximum reproducibility, the hardware and software stack was kept constant during all experiments. 
The official implementations of \gls{yolo5} (v6.2) \cite{yolov5} and \gls{mmdet} (v2.25.1) \cite{mmdetection} were used as the basis for our adaptions and experiments.
The methods were then executed using Nvidia's A6000 GPU and Intel's Xeon Silver 4210R CPU.

\subsection{Dataset \& Evaluation Metrics}
\label{subsec:experiements_datasets}
Since there is no publicly available dataset for the task, a dataset of airport fence damages was created.
Therefore, video sequences of different sections were recorded using two different camera models, namely a FLIR\footnote{\href{https://www.flir.eu/}{flir.eu}, Date: 01/09/2023} camera model and Panasonic's GH5\footnote{\href{https://www.panasonic.com/}{panasonic.com}, Date: 01/09/2023}.
A total of 5 datasets were recorded, 3 with the FLIR and 2 with the GH5 camera. Then all images with damage were labeled, images without damage were sorted out and were not considered further.
This results in 5 video sequences with an overall 475 video frames and 725 annotated damages, divided into 104 climb-over defects and 621 holes.
The images recorded with the FLIR camera have a resolution of $1920\times1200$ and those with the GH5 camera of $1920\times1080$, respectively.

\begin{table}[t]
        \centering
        \resizebox{0.7\linewidth}{!}{
        \begin{tabular}{cccc}
        \toprule
        Case & Training & Validation & Test\\
        \midrule
        1 &FLIR &FLIR &FLIR\\
        2 &FLIR &FLIR &GH5\\
        3 &FLIR+GH5 &FLIR+GH5 &FLIR+GH5\\
        \bottomrule                
        \end{tabular}
        }
        \caption{\textbf{Dataset splits} -- Each investigation case specifies which the datasets used for training, validation, and testing. Training and evaluation are performed in each examination case according to the \acrshort{loocv}.}
        \label{tab:experimentsdatasets}
        \vspace{-0.4cm}
\end{table}
This work considers three different cases, each reflecting another real-world scenario.
The cases differ regarding the training, validation, and testing data, as shown in ~\cref{tab:experimentsdatasets}.
Case 1 is the specialization case when training data from the exact camera used in the application is available.
Case 2 evaluates the generalization performance since training and test data originate from different camera models with dissimilar characteristics.
In the last Case 3, data from both camera models are used for all splits to evaluate the case when diverse data is available for training.

To ensure meaningful evaluation results, \acrfull{loocv} is performed in each of the three study cases to compensate for the small size of the dataset.
In each split, another video sequence is leveraged for training, resulting in 12 splits.

The \gls{coco} $\gls{ap}$~\cite{coco_map} serves as the primary metric for both evaluation and validation.
The results given represent the average across all three cases and will be abbreviated as Avg. \textit{\gls{ap}} in the following.

\subsection{Baseline}
\label{subsec:experiements_baseline}
Each method's baseline is evaluated on the 12 \gls{loocv} splits.
For this purpose, the original implementations of the methods were slightly modified.
For \gls{yolo5}, only Pytorch's recommended measures for reproducibility\footnote{\href{https://pytorch.org/docs/1.12/notes/randomness.html}{pytorch.org}, Date 01/09/2023} were added.
This ensures better comparability of experiments.
Unfortunately, this was impossible for the other three methods in \gls{mmdet} 2.25.1. 
Nevertheless, to reduce the standard deviation between the training runs and to be able to make more meaningful comparisons, three runs were performed for each data split.
For training, four changes were made to the original configurations.
First, the batch size was reduced from 32 to 8 to allow a training with faster gradient descent.
Second, to reduce the oscillation of the metrics validation curve during training, the learning rate was reduced to 5e-2.
Third, the number of epochs had to be doubled for training convergence. Fourth and last, FP16 built-in training for faster training and lower memory consumption is used.

For all models, pre-trained \gls{coco} models are utilized.
The models were then fine-tuned with the fence inspection dataset, whereby the resolution was adjusted to 768 pixels on the longest image side.
\cref{tab:experiments-od_methods-repro-results} provides the baseline results of the four methods.
\begin{table}[t]
    \centering
    \resizebox{\linewidth}{!}{
    \begin{tabular}{ll|c|c}
        \toprule
        \multirow{2}{*}{Method} & \multirow{2}{*}{Backbone} & Avg. & Case 2\\
        && $\gls{ap}$ & $\gls{ap}$\\
        \midrule
        \multirow{5}{*}{\gls{yolo5}~\cite{yolov5}} &
          n6 & 53.52$\pm$21 &25.86$\pm$8\\
          & s6 & 55.33$\pm$21 &27.42$\pm$7\\
          & m6 & 59.53$\pm$17  &37.44$\pm$2\\
          & l6 & 61.37$\pm$15 &41.84$\pm$4\\
         & x6 & \textbf{62.19}$\pm$14 &\textbf{43.34}$\pm$0\\
        \midrule
         \multirow{3}{*}{\gls{tood}~\cite{tood}}
          & \gls{resnet}50 &66.14$\pm$11 &50.42$\pm$2 \\
          & \gls{resnet}101&67.03$\pm$12 &\textbf{51.95}$\pm$4\\
          & ResNeXt101-64x4d &\textbf{67.10}$\pm$12 &50.84$\pm$2\\
        \midrule
        \multirow{3}{*}{\gls{vfnet}~\cite{varifocalnet}}  
         & \gls{resnet}50 &65.64$\pm$14 &47.22$\pm$3\\
         & \gls{resnet}101 &65.78$\pm$13 &47.86$\pm$2\\
         & ResNeXt101-64x4d &\textbf{67.75}$\pm$12 &\textbf{50.28}$\pm$3\\
        \midrule
        Def. \gls{detr}~\cite{deformable_detr}  
        & \gls{resnet}50   &\textbf{61.13}$\pm$14 &\textbf{42.11}$\pm$5\\
        \bottomrule
    \end{tabular}
    }
    \caption{\textbf{Baseline results} -- Different backbone configurations for each method are compared. For \gls{tood} and \gls{vfnet}, all configs use \glspl{dconv}~\cite{dcn,dcnv2} and Multi-Scaling as additional data augmentation strategy. The best result for each configuration is highlighted in bold.
    }
    \label{tab:experiments-od_methods-repro-results}
    \vspace{-0.5cm}
\end{table}

The results indicate that \gls{tood} and \gls{vfnet} provide the best results with 67.10\% and $\gls{ap}$ 67.75\% $\gls{ap}$. 
\gls{yolo5} achieves worse outcomes with 62.19\% $\gls{ap}$, though still surpassing Deformable \gls{detr} by 2.06\% points. 
One reason for the poor accuracy of Deformable \gls{detr} could be the limited training data, a general problem with transformers. 
Since the efficiency of Deformable \gls{detr} is significantly worse than \gls{yolo5} due to its transformer-based construction, the Deformable \gls{detr} method is not considered further in the remainder of this paper.
One reason for the poorer results of \gls{yolo5} is the subpar generalization capability. 
Comparing the results for Case 2 in \cref{tab:experiments-od_methods-repro-results}, it is apparent that the anchor-free \gls{tood} and \gls{vfnet} methods generalize remarkably stronger to unseen data.
Whether this weakness of \gls{yolo5} remains despite the optimizations in the further chapters is investigated in~\cref{subsec:experiements-backbone}.

\subsection{Regularization}
\label{subsec:experiements-meth_opt}
After training the baseline, optimizations are made for the three remaining methods.
We have adjusted the \gls{yolo5} implementation to enable training with rectangular images training in conjunction with random shuffling and mosaic data augmentation~\cite{yolov5}.
Furthermore, different hyperparameter settings proved beneficial for the m6, l6, and x6 variants of \gls{yolo5} to achieve better convergence toward the global optimum and prevent overfitting.
On the one hand, the OneCycle learning rate~\cite{onecyclelr} is increased from 1e-4 to 1e-3 to enable faster convergence of the deeper models m6, l6 and x6 and better exploit the hill climbing properties in gradient optimization.
Second, more data augmentation is employed for enhanced regularization. 
For this, the percentage of image scaling is increased from $[-50\%,+50\%]$ to $[-90\%,+90\%]$.
In addition, MixUp~\cite{mixup} is applied with a probability of 10\%.
Regarding \gls{tood} and \gls{vfnet}, no significant enhancements were observed.
\begin{table}[t]
    \centering
    \resizebox{0.6\linewidth}{!}{
    \begin{tabular}{l|cc|c}
        \toprule
        Backbone       &Params &\acrshort{flops} & Avg.\\
                    &$(M)$ &$(B)$ &$\gls{ap}$\\
        \midrule
        n6 &3.2 &4.7 &60.71$\pm$18  \\
        s6 &12.6 &17 &62.36$\pm$15  \\
        m6 &35.7 &50.3&64.68$\pm$14  \\
        l6 &76.8 &111.8&\textbf{66.05}$\pm$14  \\
        x6 &140.7 &210.5 &64.85$\pm$14 \\
    \bottomrule
    \end{tabular}
    }
    \caption{\textbf{\gls{yolo5} baseline optimization results} -- The best result is highlighted in bold.
    }
    \label{tab:experiments-od_methods-best_config}
\end{table}

The optimized \gls{yolo5} results are presented in \cref{tab:experiments-od_methods-best_config}.
The results significantly surpass the baseline results. 
This is attributed to the increased diversity of data during training through Mosaic Data Augmentation and further regularization against overfitting introduced by shuffling. 
In total, these adjustments resulted in an improvement of 3.86\% points in $\gls{ap}$ when comparing the best configurations. 
However, the best model is not the largest x6, but l6.
The x6 model tends to overfit and performs notably worse with 64.85\% $\gls{ap}$. 
Even the increased data augmentation and additional regularization cannot compensate for this. Therefore, \gls{yolo5}l6 is used as the best model in the following. 

\subsection{Contrast Adjustment}
\label{subsec:experiements-cont_adj}
\begin{figure}
    \centering
    \subfloat[Original]{
        \label{fig:data-original-clahe-histequ-org_2}
        \includegraphics[width=0.3\linewidth]{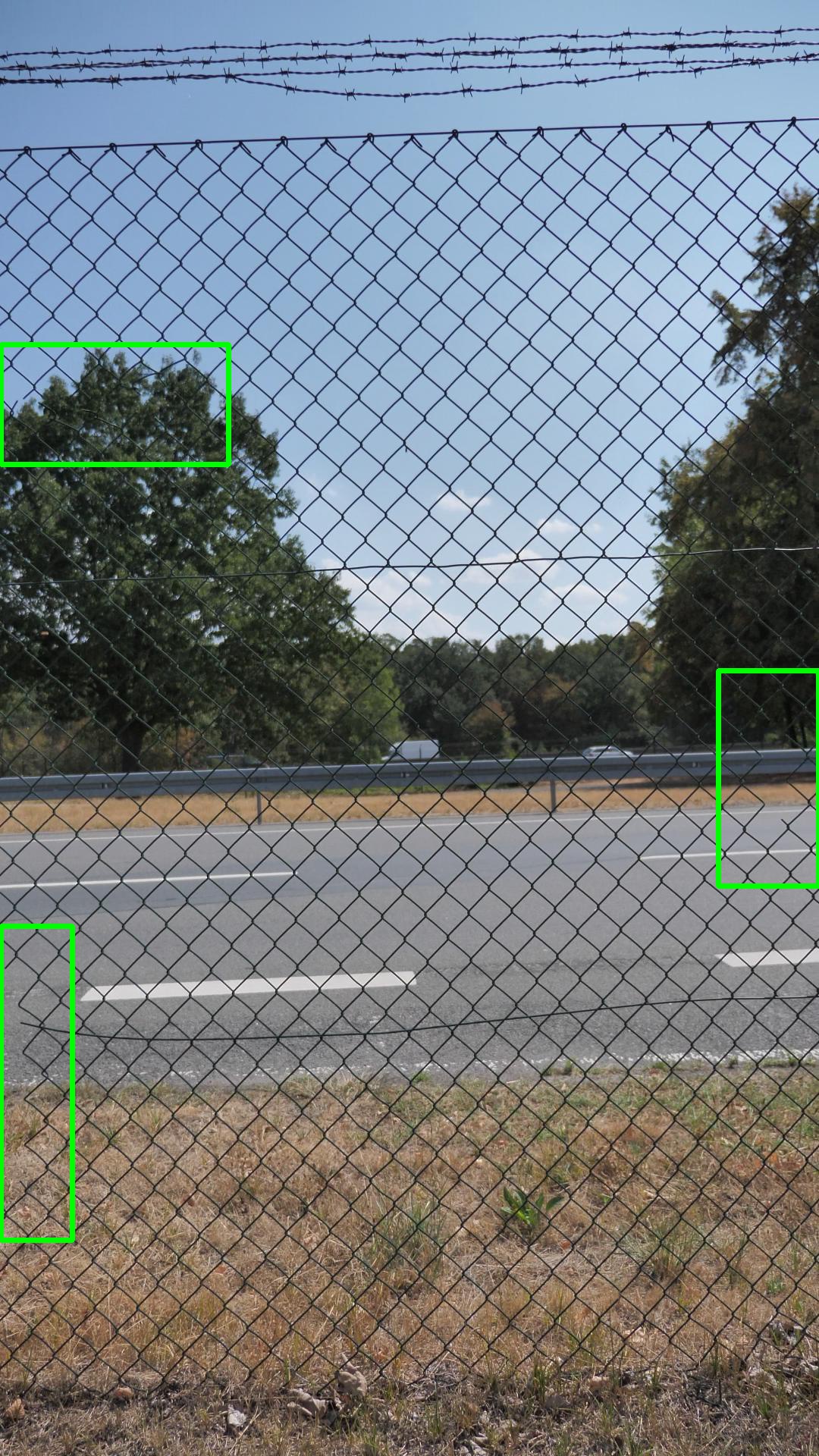}
    }
    \subfloat[\gls{clahe}]{
         \label{fig:data-original-clahe-histequ-clahe_2}
        \includegraphics[width=0.3\linewidth]{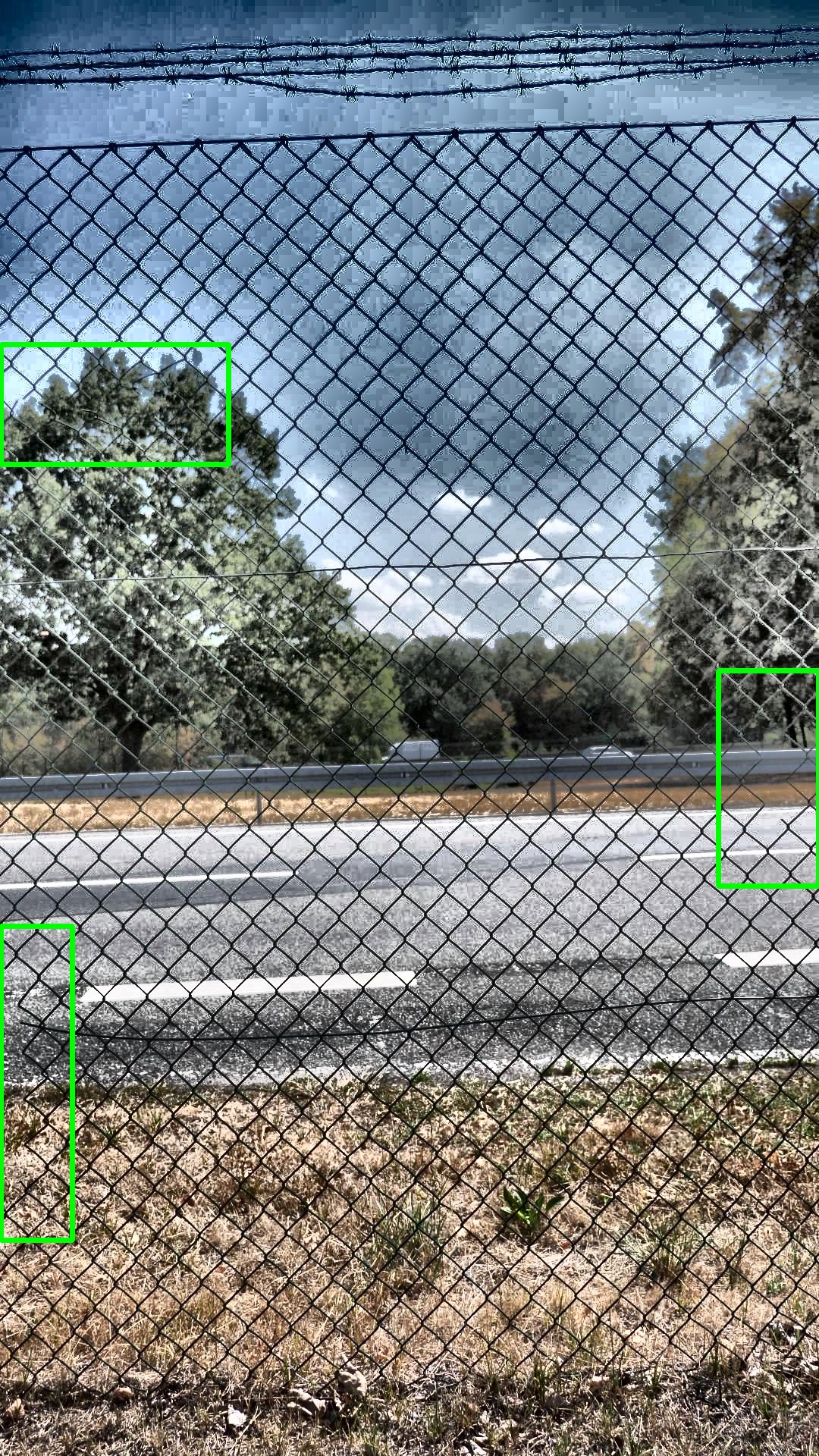}
    }
    \subfloat[\gls{histequ}]{
         \label{fig:data-original-clahe-histequ-histequ_2}
        \includegraphics[width=0.3\linewidth]{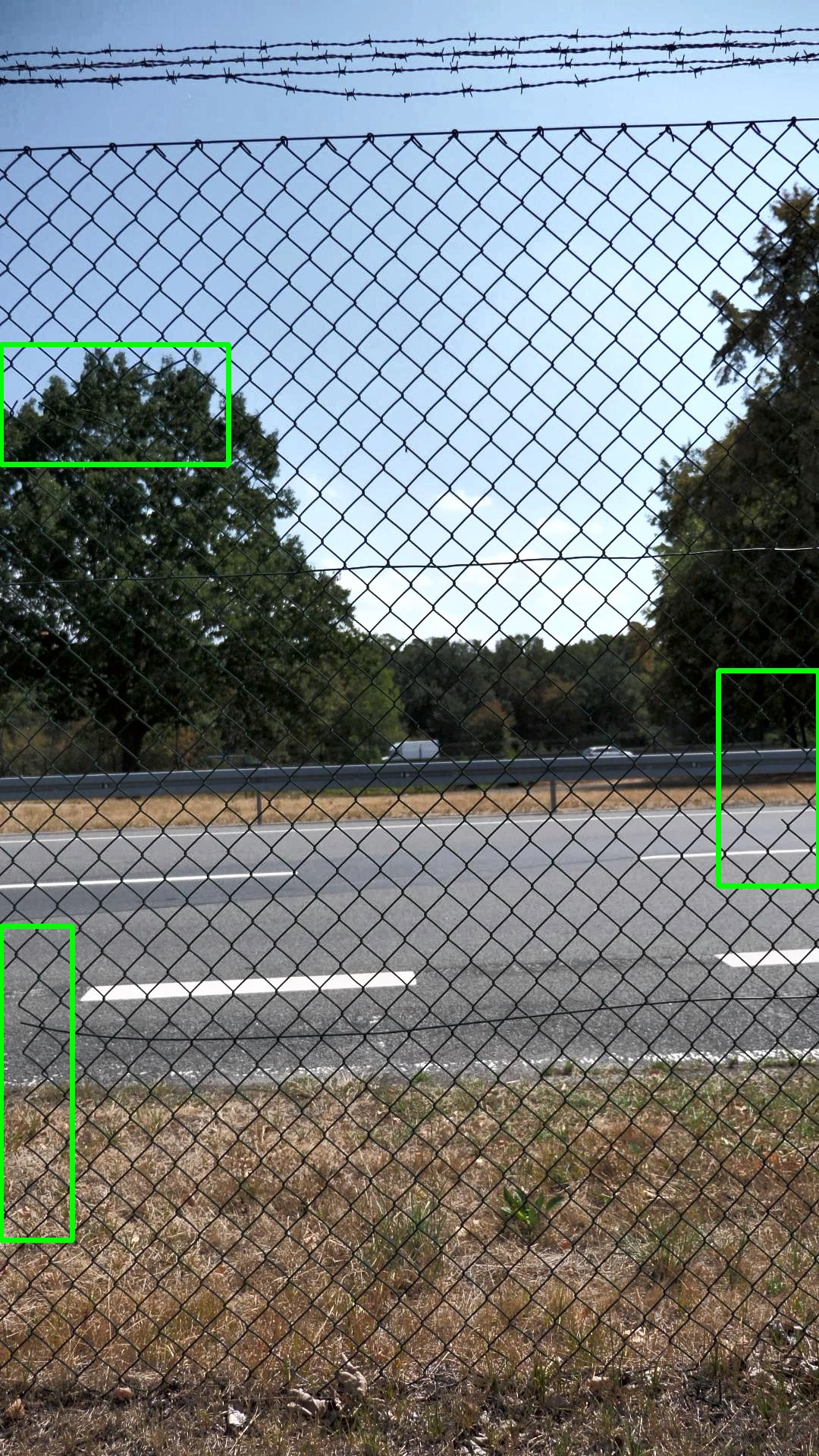}
    }
    \caption{\textbf{\gls{clahe} vs. \gls{histequ}} -- \gls{clahe} leads to over-adjustments compared to \gls{histequ}. Due to the good contrast in the original image, the contrast is lowered by \gls{histequ}. Nevertheless, the holes are clearly recognizable. In contrast, \gls{clahe} results in too bright areas. Similar to dark areas, the fence structure is difficult to recognize.
    }
    \label{fig:experiements-cont_adj}
    \vspace{-0.5cm}
\end{figure}

\begin{table}[t]
    \centering
    \resizebox{\linewidth}{!}{
    \begin{tabular}{ll|c|ccc}
            \toprule
            \multirow{2}{*}{Method}       & \multirow{2}{*}{Experiment}     & Avg.       & Case 1   & Case 2   & Case 3   \\
                        &               & $\gls{ap}$ &$\gls{ap}$      &$\gls{ap}$      &$\gls{ap}$     \\
            \midrule
            \multirow{3}{*}{\gls{yolo5}~\cite{yolov5}} & Regularization &66.05$\pm$14 &73.45$\pm$4 &47.74$\pm$4 &76.97$\pm$2 \\
            &\gls{clahe}  &66.22$\pm$14 &73.87$\pm$3 &47.28$\pm$1 &\textbf{77.51}$\pm$2 \\
            &\gls{histequ}  &\textbf{67.16}$\pm$14 &\textbf{75.46}$\pm$4 &\textbf{48.88}$\pm$2 &77.14$\pm$2 \\
            \midrule
            \multirow{3}{*}{\gls{tood}~\cite{tood}} & Baseline &67.10$\pm$12 &73.24$\pm$4 &50.84$\pm$2 &\textbf{77.24}$\pm$2 \\
            &\gls{clahe} &64.52$\pm$14 &72.07$\pm$4 &45.86$\pm$5 &75.63$\pm$3 \\
            &\gls{histequ} &\textbf{67.62}$\pm$11 &\textbf{73.33}$\pm$4 &\textbf{52.31}$\pm$1 &77.22$\pm$2 \\
            \midrule
            \multirow{3}{*}{\gls{vfnet}~\cite{varifocalnet}} & Baseline &\textbf{67.75}$\pm$12 &\textbf{74.14}$\pm$2 &\textbf{50.87}$\pm$3 &78.25$\pm$2 \\
             &\gls{clahe} & 65.14$\pm$15 &72.97$\pm$3 &44.54$\pm$3 &77.91$\pm$2 \\
             &\gls{histequ} &\textbf{67.49}$\pm$13 &\textbf{73.73}$\pm$2 &\textbf{50.40}$\pm$3  &\textbf{78.36}$\pm$2 \\
            \bottomrule
        \end{tabular}
        
    }
    \caption{\textbf{\gls{histequ} and \gls{clahe} results} -- Results obtained with the best configuration of methods. The first line of each block indicates the best experiments so far on the original dataset. For comparison, the best results of \gls{yolo5} were taken from \cref{subsec:experiements-meth_opt} and for \gls{tood} and \gls{vfnet} from~\cref{subsec:experiements_baseline}. The best results for each \gls{dl} method are highlighted in bold.}
    \label{tab:experiments-od_methods-contrastadjustment-pp}
    \vspace{-0.5cm}
\end{table}
The two contrast adjustment methods \gls{clahe} and \gls{histequ} are compared in~\cref{tab:experiments-od_methods-contrastadjustment-pp}.
The results indicate superior performance of the global method \gls{histequ} regardless of the detection approach. 
One reason for this could be the over-adjustment of \gls{clahe} in certain regions.
Especially worse results concerning the generalization Case 2 support this hypothesis. 
Since GH5 images already show good contrast, an additional contrast adjustment leads to over-adjustment.
\cref{fig:experiements-cont_adj} visualizes the differences between both methods for an image captured by the GH5 camera.
The \gls{clahe} method, as shown in~\cref{fig:data-original-clahe-histequ-clahe_2}, clearly over-adjusts, compared to \gls{histequ}, which is depicted in~\cref{fig:data-original-clahe-histequ-histequ_2}.
These overfits occur in areas with a high difference between light and dark pixels, such as trees and the sky.
This leads to a very unnatural appearance of the image.
As a result, parts of the fence structure are hardly recognizable.

\subsection{Hyperparameter Optimization}
\label{subsec:experiements-hyp_opt}
The hyperparameters are optimized using \gls{histequ} preprocessing.
Analogous to regularization, the \gls{mmdet} implementation methods \gls{tood} and \gls{vfnet} provide no significantly improved results.
As a result, hyperparameter optimization focuses on \gls{yolo5}.
We found that choosing a learning rate of 5e-3 and applying image weighting turned out to be beneficial.
This manual hyperparameter optimization increases the $\gls{ap}$ from 67.16\% to 68.45\%.
Besides, numerous settings freezing different stages of the backbone, and the use of Adam~\cite{adam} and AdamW~\cite{adamW} as the optimizer were evaluated to achieve stronger regularization and thereby a more stable training. We also evaluated several settings regarding the affine transformations to achieve a higher generalization.
However, none of the mentioned adjustments led to significantly improved results.

Thereafter, an automatic hyperparameter tuning was performed.
First, all previous internal evaluations of all 12 \gls{loocv} splits were used, and the Pearson Correlation Coefficient between the average $\gls{ap}$ across the splits and the $\gls{ap}$ of each individual split was determined.
Subsequently, the split is identified that correlates most with the average $\gls{ap}$ over all splits.
This split is leveraged for automatic hyperparameter tuning. 

We apply the \gls{ga}~\cite{genetic_algo_ai} implemented in \gls{yolo5} for automatic hyperparameter optimization in the predefined configuration, except for a few changes.
Based on our previous findings, we reduce the defined search space and exclude the affine transformations rotation, shearing, perspective, and flipping since their use leads to significant degradation. 
Finally, automatic hyperparameter tuning is executed for 500 iterations with the remaining 21 hyperparameters.
In each iteration, one or more hyperparameter adjustments are sampled according to the \gls{ga} policy and then evaluated in a complete training run without early stopping.
The most significant effects were observed in reducing the probability of Mosaic Data Augmentation from 100\% to 91.5\%, since the network requires original data to capture the inherent structure. 
Additionally, increasing the variation of the saturation in ColorJitter augmentation from $[-70\%, +70\%]$ to $[-89\%, +89\%]$ lead to notable improvement.
In total, the optimized model achieves 69.09\% in $\gls{ap}$ on average across all data splits.

\begin{table}[t]
    \centering
    \resizebox{\linewidth}{!}{
    \begin{tabular}{llc|cc|c|c}
        \toprule
        Method    & Backbone &\gls{dconv}  &Param & \acrshort{flops} & \gls{coco}   & Avg. \\
                    &       &                  &$(M)$    &$(B)$  & $\gls{ap}$ &$\gls{ap}$     \\
        \midrule
        \gls{tood} &ResNet101 &$\times$ &53.2 &149.0 &49.3 &67.62$\pm$11\\
        \gls{tood}\textsuperscript{+} &ConvNeXt-T &$\times$ & 35.7& 154.2&44.9&65.86$\pm$13\\
        \gls{tood} &ConvNeXt-T &$\times$ & 35.7& 154.2& 48.6 &\textbf{67.76}$\pm$12\\
        \gls{tood} &Res2Net101 &$\times$ &51.7&220.1&45.2&62.05$\pm$9 \\
        \gls{tood} &Res2Net101 &$\checkmark$ &54.5&187.4&50.9&65.13$\pm$9 \\
        \midrule
        \gls{vfnet} &ResNeXt101-32x4d &$\checkmark$ &55.1 &208.1 &49.7 &67.49$\pm$13\\
        \gls{vfnet}\textsuperscript{+} &ConvNeXt-T &$\times$ &36.5&161.1&44.5&66.63$\pm$13\\
        \gls{vfnet} &ConvNeXt-T &$\times$ &36.5&161.1&48.9&64.14$\pm$15 \\
        \gls{vfnet}\textsuperscript{*} & Res2Net101 &$\times$ &52.4&227.0&49.2&65.77$\pm$14 \\
        \gls{vfnet}\textsuperscript{*} & Res2Net101 &$\checkmark$ &54.9&187.5&51.1&\textbf{68.09}$\pm$12 \\
        \bottomrule
    \end{tabular}
    }
    \caption{\textbf{\gls{tood} and \gls{vfnet} results with new backbones} -- In each case, the first line of a block represents the best training so far of the methods from~\cref{tab:experiments-od_methods-contrastadjustment-pp}. 
    Best Avg. $\gls{ap}$ (calculated on our fence dataset) is marked bold.
    $^{*}$Pre-trained weights used.
    $^{+}$Uses original configurations.}
    \label{tab:experiments-backbone-coco_results}
    \vspace{-0.5cm}
\end{table}

\subsection{Backbones}
\label{subsec:experiements-backbone}
After hyperparameter tuning, modern \gls{sota} backbones are evaluated in conjunction with \gls{tood} and \gls{vfnet}.
Besides, the influence of using \gls{dconv} within the Res2Net architecture is examined.
The results of the so-far best models and the new pre-trained ones are given in~\cref{tab:experiments-backbone-coco_results}. For each training session, the $\gls{ap}$ of the pre-trained network on the \gls{coco} dataset is presented in addition to the $\gls{ap}$ for our dataset.
In the case of \gls{tood}, for instance, the best pre-trained network on \gls{coco} is not necessarily the best network on our dataset.
This is because the classes and the class semantics in the \gls{coco} dataset deviate considerably from those in this work.
However, it provides a rough indication when further consideration of a backbone is not promising.
The findings indicate that \gls{tood} in conjunction with ConvNeXt achieves the highest accuracy.
Regarding \gls{vfnet}, Res2Net as the backbone performs best.
Despite the significant improvement in accuracy with the new backbones, \gls{tood} and \gls{vfnet} do not surpass \gls{yolo5} in $\gls{ap}$. 
Since \gls{yolo5} is also more resource efficient due to its design as a real-time object detector, \gls{yolo5} was selected as the best model and is utilized in the remainder of this paper.

\subsection{In-depth Analysis}
\label{subsec:experiements-further_analysis}
So far, all analyses have been performed with the $\gls{ap}$ across all types of failure.
This showcased remarkable progress over the baseline with 6.9\% points.
This section thoroughly delves into the effects of the proposed optimizations to identify strengths and weaknesses of the system.

\begin{table}
    \centering
    \resizebox*{\linewidth}{!}{
    \begin{tabular}{l|l|c|c|c}
        \toprule
        \multirow{2}{*}{Damage Type} & \multirow{2}{*}{Metric} &\multicolumn{2}{c|}{\gls{yolo5}} &\multirow{2}{*}{Improvement}\\
        & & Baseline &Hyp. Opt. & \\
        \midrule
        \multirow{4}{*}{All} &$\gls{ap}$ &$62.19\pm14$ &$69.09\pm12$ &\textcolor{KITgreen}{+6.90}\\
        & $\gls{ap}^{small}$  &$21.69\pm14$ &$26.80\pm18$ &\textcolor{KITgreen}{+5.11}\\
        & $\gls{ap}^{medium}$ &$65.04\pm11$ &$70.75\pm9$ &\textcolor{KITgreen}{+5.71}\\
        & $\gls{ap}^{large}$  &$68.52\pm25$ &$83.41\pm10$ &\textcolor{KITgreen}{+14.89}\\
        \midrule
        \multirow{4}{*}{Climb over defect} & $\gls{ap}$ &$77.12\pm12$ &$86.53\pm6$ &\textcolor{KITgreen}{+9.41}\\
        & $\gls{ap}^{small}$  &$-$ &$-$ &$-$\\
        & $\gls{ap}^{medium}$ &$80.80\pm10$ &$89.50\pm4$ &\textcolor{KITgreen}{+8.70}\\
        & $\gls{ap}^{large}$  &$77.81\pm12$ &$86.80\pm6$ &\textcolor{KITgreen}{+8.99}\\
        \midrule
        \multirow{4}{*}{Hole} & $\gls{ap}$ &$47.26\pm18$ &$51.66\pm18$ &\textcolor{KITgreen}{+4.40}\\
        & $\gls{ap}^{small}$  &$21.69\pm14$ &$26.80\pm18$ &\textcolor{KITgreen}{+5.11}\\
        & $\gls{ap}^{medium}$ &$50.90\pm17$ &$54.30\pm18$ &\textcolor{KITgreen}{+3.40}\\
        & $\gls{ap}^{large}$  &$45.82\pm41$ &$74.88\pm16$ &\textcolor{KITgreen}{+29.06}\\
        \bottomrule
    \end{tabular}
    }
    \caption{\textbf{Defect results} -- Comparison between the different types and sizes of damages. Results are given for baseline (see~\cref{subsec:experiements_baseline}) and the Hyp. Opt. (see~\cref{subsec:experiements-hyp_opt}) as the best training of \gls{yolo5}. The different ranges small, medium and large were defined as follows: $0 < \gls{ap}^{small} \leq$ 24,000 pixels, 24,000 \text{pixels} $< \gls{ap}^{medium} \leq$ 100,000 pixels and 100,000 pixels $< \gls{ap}^{large}$. 
    }
    \label{tab:experiments-od_methods-fencetypes}
    \vspace{-0.5cm}
\end{table}
\noindent\textbf{Types and area size of fence defects:}
\cref{tab:experiments-od_methods-fencetypes} investigates the results for each defect type and different sizes of damages for \gls{yolo5}.
For this purpose, the damages are divided into three classes based on the covered area in pixels. 
Damage up to a size of 24,000 pixels is considered small.
Correspondingly, damage ranging from 24,000 pixels to 100,000 pixels and over 100,000 pixels as medium or large, respectively.
Thereby, 8\% of all damages are small, 77\% medium and 15\% small.
In general, the $\gls{ap}$ difference between the damage types decreases by the optimizations. 
However, the difference is still a considerable 24.87\% points.
The stronger detection of the climb-over-protection defects can be explained by their characteristic appearance and by the angle of view.
Typically, the damage is in front of the bright sky and, therefore, discriminates well from the background, even under poor lighting conditions (see~\cref{fig:intro_example_dataset}).
In contrast, the wire mesh exhibits poor contrast. The next striking feature in the baseline is the very high standard deviation of 41 for large holes. This finding suggests unstable generalization capabilities and great dependence from the training and validation data.
One reason for this is that in the $\gls{ap}^{large}$, the holes are nearly normally distributed up to 500,000 pixels.
Therefore, training splits with few large boxes may exceed the generalization capability of the baseline to evaluation splits with huge boxes.
The results for the optimized hyperparameters suggest greatly improved generalization capabilities.
This improvement contributes to better results over all damages.
The detection accuracy for the different damage sizes consistently shows the expected behavior that larger objects are detected more accurately than smaller objects.
However, the difference in accuracy is very large in some cases. 
For instance, the difference for the best model between $AP^{small}$ and $AP^{medium}$ is 35.14\% points.
Even with a good contrast ratio, small holes caused by, e.g., minor cracks, are difficult to separate from sound parts of the mesh.
Interestingly, medium-sized climb over defects are detected more robustly than large ones, regardless of the approach.
This is due to a lack of training data depicting large climb over defects.
In general, it can be concluded that climb over defects are easier to localize due to their position and larger size. 
In total, a 34.87\% points difference in $\gls{ap}$ between such damages and holes is observed for the best model.

\begin{table}
    \centering
    \resizebox*{\linewidth}{!}{
    \begin{tabular}{l|c|c|c}
        \toprule
        \multirow{2}{*}{Metric} & \multicolumn{2}{c|}{\gls{yolo5}} & \multirow{2}{*}{Improvement}\\
        & Baseline & Hyp. Opt. & \\
        \midrule
        Class Error &$0.44\pm1$ &$0.10\pm0$ &\textcolor{KITgreen}{$-0.33$}\\
        Localization Error  &$3.03\pm3$ &$0.80\pm1$ &\textcolor{KITgreen}{$-2.23$}\\
 As+Localization Error &$0.03\pm0$ &$0\pm0$ &\textcolor{KITgreen}{$-0.03$}\\
        Duplicate Error &$0.33\pm0$ &$0.26\pm0$ &\textcolor{KITgreen}{$-0.07$}\\
        Background Error &$0.82\pm1$ &$1.55\pm2$ &\textcolor{KITred}{$+0.73$}\\
        Missing Error &$5.78\pm5$ &$1.44\pm1$&\textcolor{KITgreen}{$-4.34$}\\
        \midrule
        \gls{fp} Rate &$3.79\pm3$ &$4.06\pm4$ &\textcolor{KITred}{$+0.27$}\\
        \gls{fn} Rate &$7.83\pm6$ &$3.36\pm3$ &\textcolor{KITgreen}{$-4.47$}\\
        \bottomrule
    \end{tabular}
    }
    \caption{\textbf{\gls{yolo5} analysis} -- Comparison of \gls{yolo5} baseline and optimized results. Metrics are calculated with the \gls{tide} library.
    }
    \label{tab:experiments-od_methods-tide}
\end{table}

\begin{table}
    \centering
    \resizebox*{0.75\linewidth}{!}{
    \begin{tabular}{l|c|c|c}
        \toprule
        \multirow{2}{*}{Metric} & \multicolumn{2}{c|}{\gls{yolo5}} & \multirow{2}{*}{Improvement}\\
        & Baseline & Hyp. Opt. & \\
        \midrule
        $\gls{ap}^{50}$ &$87.52\pm10$ &$91.73\pm8$ &\textcolor{KITgreen}{$+4.21$}\\
        $\gls{ap}^{55}$ &$86.47\pm11$ &$90.17\pm8$ &\textcolor{KITgreen}{$+3.60$}\\
        $\gls{ap}^{60}$ &$82.47\pm11$ &$87.23\pm9$ &\textcolor{KITgreen}{$+4.76$}\\
        $\gls{ap}^{65}$ &$77.68\pm15$ &$82.79\pm11$ &\textcolor{KITgreen}{$+5.11$}\\
        $\gls{ap}^{70}$ &$72.22\pm18$ &$76.94\pm14$ &\textcolor{KITgreen}{$+4.72$}\\
        $\gls{ap}^{75}$ &$65.52\pm19$ &$71.27\pm15$ &\textcolor{KITgreen}{$+5.75$}\\
        $\gls{ap}^{80}$ &$58.35\pm18$ &$65.77\pm16$ &\textcolor{KITgreen}{$+7.42$}\\
        $\gls{ap}^{85}$ &$49.41\pm19$ &$59.39\pm15$ &\textcolor{KITgreen}{$+9.98$}\\
        $\gls{ap}^{90}$ &$34.01\pm16$ &$45.66\pm13$ &\textcolor{KITgreen}{$+11.65$}\\
        $\gls{ap}^{95}$ &$8.29\pm7$ &$20.15\pm8$ &\textcolor{KITgreen}{$+11.86$}\\
        \bottomrule
    \end{tabular}
    }
    \caption{\textbf{Influence of \acrshort{iou} threshold} -- Comparison of \gls{yolo5} $\glspl{ap}$ for different \acrshort{iou} thresholds. 
    }
    \label{tab:experiments-od_methods-ap_ious}
    \vspace{-0.5cm}
\end{table}
\noindent\textbf{Error sources:}
So far, the analysis has been conducted quantitatively via the $\gls{ap}$. 
In this section, the \gls{tide}~\cite{tidecv} library is utilized to break down the error sources.
For this purpose, different error types are presented in the upper part of~\cref{tab:experiments-od_methods-tide}.
The error types describe erroneous relationships of GT \gls{bb} and predicted \gls{bb}, such as a deviating position or even a missed prediction.
The localization and the missing of damages have improved more than average.
The improved ability to localize damages may lead to enhanced generalization to other fence types or transfer the learned features to new contexts.
The significant reduction in localization error is due to increased $\gls{ap}$ all \glspl{iou}. The $\gls{ap}$ results with different \glspl{iou}, i.e., varying degrees of overlap with the ground truth \glspl{bb}, are shown in~\cref{tab:experiments-od_methods-ap_ious}. 
Thus, for \gls{iou} of 0.90 and 0.95 in each case over 11\% improvement was obtained. 
However, in the context of this work, the improvement of the missing damages is more relevant.
Exact recognition is not directly necessary, but can of course help with generalization.
Although the false positive rate increased slightly by 0.27\% points compared to the baseline, it is still at a low of 4.06\%. This means there would not be too many false alarms in real-world use. In principle, it is better to detect a few too many holes, which can be rechecked digitally, than to completely forget holes. The latter would jeopardize the airport's approval.
The significant improvement in missing damage is accompanied by a decrease in \gls{fn} rate. 
This has improved by 4.47\% points, implicating enhanced usefulness of the model for real-world fence inspection.

\begin{table}[t]
    \centering
    \resizebox*{\linewidth}{!}{
    \begin{tabular}{c|c|c|ccc}
        \toprule
        \multirow{2}{*}{ID} & Image Resolution & Avg.       & Case 1   & Case 2   & Case 3   \\
        & $(pixels)$ & $\gls{ap}$ & $\gls{ap}$      & $\gls{ap}$      & $\gls{ap}$ \\
        \midrule
        R1 & 288$\times$384 &57.2$\pm$19 &68.00$\pm$4 &31.72$\pm$7 &71.88$\pm$2\\
        R2 & 320$\times$512 &65.16$\pm$12 &71.44$\pm$2 &48.88$\pm$1 &75.16$\pm$2 \\
       R3 & 416$\times$640 &67.26$\pm$11 &72.61$\pm$2 &52.34$\pm$2 &76.83$\pm$2 \\
       \midrule
       R4 & 512$\times$768 &69.09$\pm$12 &74.64$\pm$3 &53.82$\pm$2 &78.81$\pm$2 \\
       \midrule
        R5 & 624$\times$960 &70.78$\pm$9 &74.96$\pm$2 &\textbf{58.63}$\pm$2 &78.75$\pm$1 \\
         R6 & 736$\times$1152 &70.86$\pm$10 &75.49$\pm$1 &58.09$\pm$2 &78.99$\pm$1 \\
         R7 & 848$\times$1344 &\textbf{70.98}$\pm$11 &\textbf{76.83}$\pm$1 &56.21$\pm$1 &79.91$\pm$1 \\
         R8 & 960$\times$1536 &70.64$\pm$11 &76.44$\pm$3 &55.62$\pm$2 &\textbf{79.87}$\pm$1 \\
         R9 & 1136$\times$1728 &69.86$\pm$10 &75.31$\pm$2 &56.03$\pm$0 &78.25$\pm$1 \\
         R10 & 1248$\times$1920 &69.86$\pm$12 &75.62$\pm$3 &54.58$\pm$4 &79.39$\pm$1 \\
        \bottomrule
    \end{tabular}
    }
    \caption{\textbf{Influence of image resolution} -- Results of experiments image resolution with the \gls{histequ} dataset and \gls{yolo5}. R4 was used in the previous experiments.}
    \label{tab:experiments-od_methods-image_resolution}
    \vspace{-0.2cm}
\end{table}
\subsection{Image Resolution}
\label{subsec:experiements-img_res}
Previous experiments have been carried out with a fixed spatial resolution of input images.
However, higher resolution imagery provides more details, which may be beneficial to the task.
The results from various resolutions are presented in~\cref{tab:experiments-od_methods-image_resolution}.
One can observe that the $\gls{ap}$ increases the larger the images but drops again when the image is larger than 848$\times$1344 pixels (R7).
The drop is due to the pre-training with the \gls{coco} dataset in a resolution of $1280\times1280$, which expects objects to have a specific size.


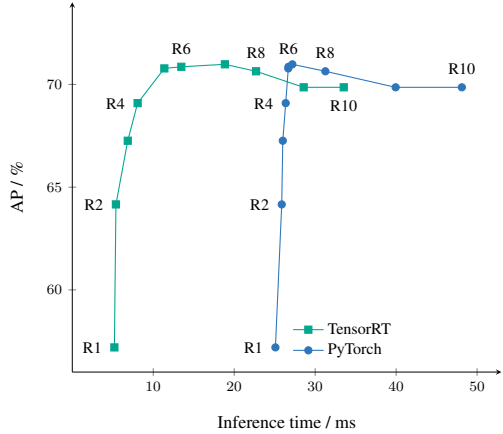
\begin{figure}[t]

    \centering
    \resizebox{0.8\linewidth}{!}{%
    \begin{tikzpicture}
        \begin{axis}[
            xmin=0,xmax=53,
            ymin=56,ymax=74,
            x label style={at={(axis description cs:0.5,-0.1)},anchor=north},
            y label style={at={(axis description cs:-0.1,.5)},rotate=90,anchor=south},
            xlabel=Inference time / ms,
            ylabel=\gls{ap} / \%,
            legend style={at={(0.5,0.02)},anchor=south west}
            ]
            \addplot[
				color=KITgreen,
				mark=square*,
				]
				table [x=trt, y=ap, col sep=comma] {data.csv};
            \node[label=left:\small R1] at (5.20,57.2) {};
            \node[label=left:\small R2] at (5.39,64.16) {};
            \node[label=left:\small R4] at (8.07,69.09) {};
            \node[label=above:\small R6] at (13.47,70.86) {};
            \node[label=above:\small R8] at (22.7,70.64) {};
            \node[label=below:\small R10] at (33.53,69.86) {};
            \addlegendentry{TensorRT}
            
            \addplot[color=KITblue,mark=*] table [x=torch, y=ap, col sep=comma] {data.csv};
            \node[label=left:\small R1] at (25.09,57.2) {};
            \node[label=left:\small R2] at (25.87,64.16) {};
            \node[label=left:\small R4] at (26.36,69.09) {};
            \node[label=above:\small R6] at (26.71,70.86) {};
            \node[label=above:\small R8] at (31.26,70.64) {};
            \node[label=above:\small R10] at (48.12,69.86) {};
            \addlegendentry{PyTorch}
        \end{axis}
    \end{tikzpicture}
    }
    \caption{\textbf{Inference time} -- Comparison between inference time and \gls{ap} results for varying image resolutions. R1, R2, etc. refer to the ID in \cref{tab:experiments-od_methods-image_resolution}. By using TensorRT, all resolutions except R10 are real-time capable. Also a significant acceleration of up to 20ms could be achieved by TensorRT. }
    \label{fig:inference}
    \vspace{-0.2cm}
\end{figure}

\subsection{Inference time}
\label{subsec:experiements-inf_time}
For the use of the model on, e.g., mobile robots, it is important to achieve a favorable tradeoff between accuracy and computation time.
\cref{fig:inference} compares the inference times of our best \gls{yolo5} model for different resolutions and with and without the use of TensorRT~\footnote{https://developer.nvidia.com/tensorrt/, Date: 01/09/2023} acceleration.
The closer the result to the top left corner, the better.
Obviously, TensorRT clearly outperforms PyTorch in inference time.
The best tradeoff between speed and accuracy is found for resolution R5 with $624\times960$ pixels.
Afterward, inference time increases and accuracy decreases.
Noticeably, up to resolution R7, approximately the same inference time is needed with PyTorch. 
This suggests a computational bottleneck outside the GPU.

\subsection{Generalization}
\label{subsec:experiements-gen}

   \begin{figure}[t]
        \centering
        \subfloat[\href{https://spectator.sme.sk/c/20602577/plane-spotters-get-observation-holes-in-the-fence-around-bratislava-airport.html}{spectator.sme.sk}]{
            \centering
            \includegraphics[height=2.5cm]{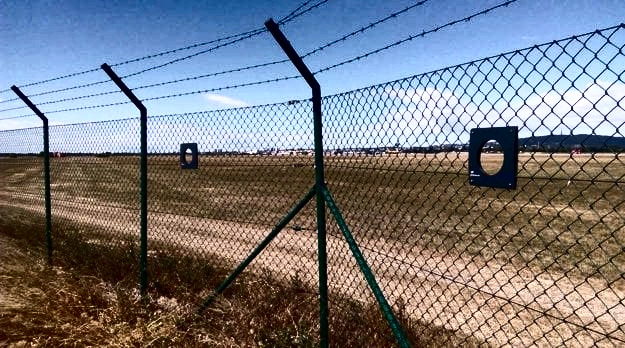}
             \label{fig:external_fence_images_yolov5-1}
        }
        \subfloat[\href{https://commons.wikimedia.org/wiki/File:Hole_in_wire_fence_2016.jpg}{wikimedia.org}]{
            \centering
            \includegraphics[height=2.5cm]{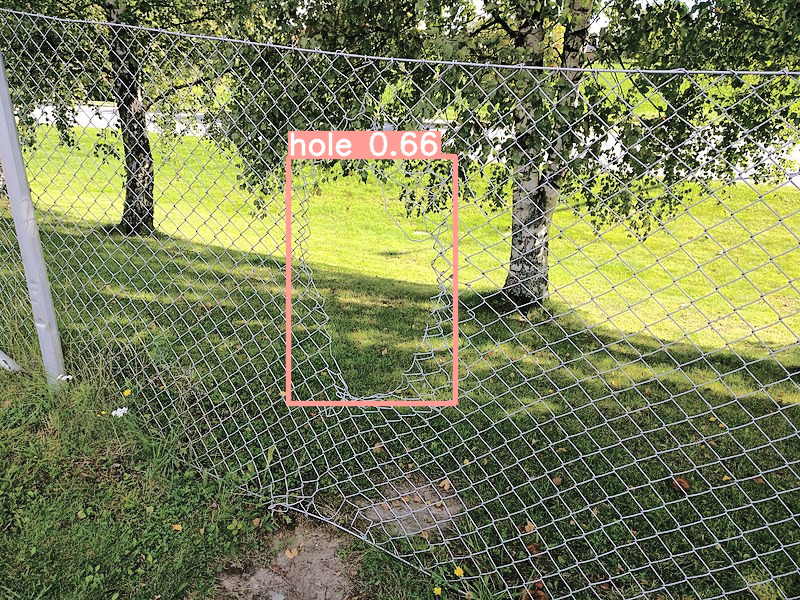}
        \label{fig:external_fence_images_yolov5-3}
        }
        \\
        \subfloat[\href{https://www.bild.de/regional/berlin/berlin-aktuell/blockade-am-airport-klima-kleber-auf-dem-rollfeld-am-berliner-flughafen-82052016.bild.html}{bild.de}]{
            \centering
            \includegraphics[height=2.5cm]{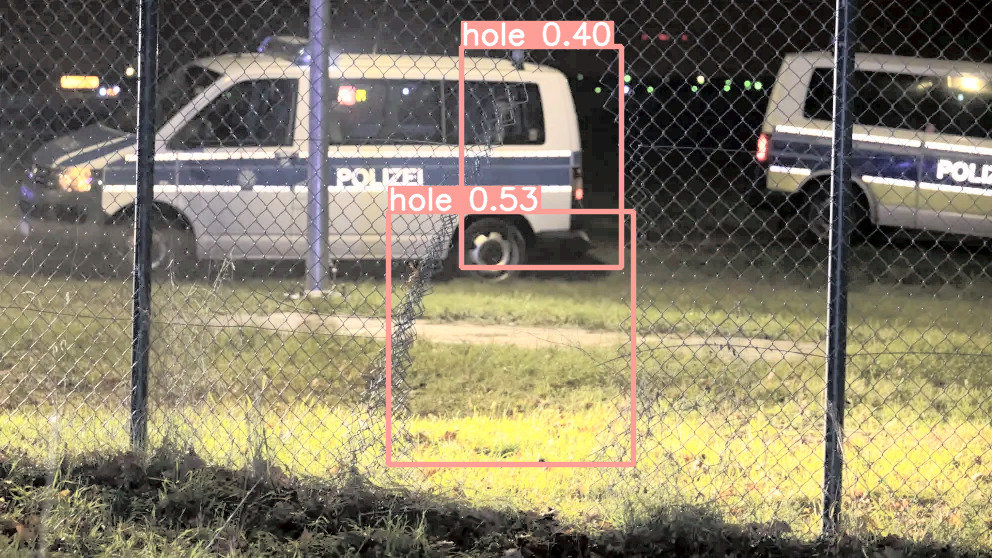}
             \label{fig:external_fence_images_yolov5-2}
        }
        \hfill
         \subfloat[\href{https://taz.de/Aktion-der-Letzten-Generation/!5898209/}{taz.de}]{
            \centering
            \includegraphics[height=2.5cm]{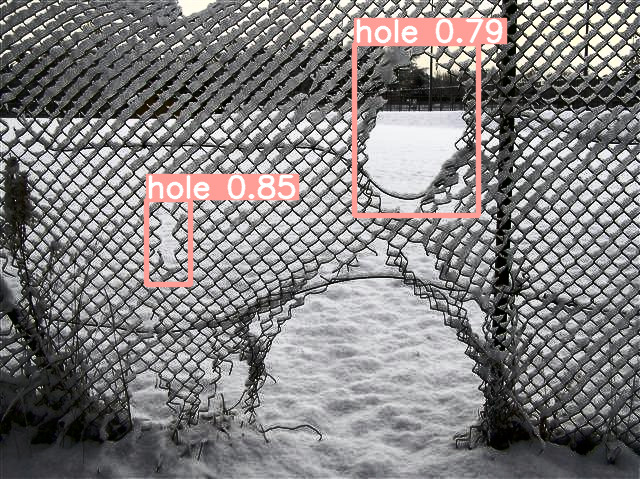}
             \label{fig:external_fence_images_yolov5-5}
        }
        \hfill
        \subfloat[\href{https://www.geograph.org.uk/photo/1626653}{geograph.org.uk}]{
            \centering
            \includegraphics[height=2.5cm]{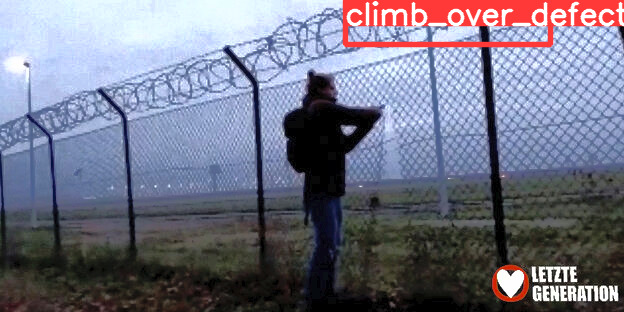}
             \label{fig:external_fence_images_yolov5-4}
        }
        \caption{\textbf{Generalization} -- \gls{yolo5} generalization results on external fence images. 
        }
        
        \label{fig:external_fence_images_yolov5}
        \vspace{-0.5cm}
    \end{figure}
As a last step, we evaluate the transferability of our model to further fences, camera models, and weather conditions to identify the strengths and also directions for future research. 
For this purpose, it is applied to external, freely available images of airport fences.
Results are visualized in~\cref{fig:external_fence_images_yolov5}.
As shown in the figure, not all fences have damage. 
For example, in \cref{fig:external_fence_images_yolov5-1}, new modules were added to the fences to facilitate photographing through the fences and avoid plane spotters from cutting holes in the fences. 
Our method does not detect these holes as damages, i.e., it works correctly.
Large holes, which are bigger than those included in the dataset, are also correctly detected, as shown in \cref{fig:external_fence_images_yolov5-2} and \cref{fig:external_fence_images_yolov5-3}.
Two holes are recognized instead of one in \cref{fig:external_fence_images_yolov5-2}.
However, this is no issue in real-world applications, as only the occurrence of damage in a specific location is relevant.
In contrast to the aforementioned examples, the hole depicted in \cref{fig:external_fence_images_yolov5-4} has a different shape and, thus, is not detected by our approach.
Future works might consider more variation regarding the shapes of holes included in the training dataset.
Furthermore, only two out of three damages are detected in the snowy environment visualized in \cref{fig:external_fence_images_yolov5-5}.
All in all, it can be concluded that the model achieves strong generalization performance to novel image sources.
However, training data with increased diversity concerning the shape of damages and weather conditions is required to address the existing weaknesses.

    

\section{Conclusion}
\label{sec:conclusion}
Within the scope of the work, the four \gls{dl} methods \gls{yolo5}, \gls{tood}, \gls{vfnet} and Deformable \gls{detr} were compared to investigate, as a first publication ever, new design rules for airport fence inspection on a small dataset. 
In conclusion, Deformable \gls{detr} as a transformer-based model does not offer any value due to the too-low data volume and the significantly lower accuracy. \gls{tood} and \gls{vfnet} could achieve higher accuracy with modern \gls{sota} backbones like ConvNeXt and Res2Net, but could not reach the accuracy and the efficiency of \gls{yolo5}.
Furthermore, we could show that YOLOv5 also provides good generalization capability on external data.

To improve the accuracy of fence analysis, it would be beneficial to separate the fence from the surrounding context. 
Although labeling such fine structures is time-consuming, recording with stereo or RGB-D cameras can provide additional information to separate the fence structure from the background.
Additionally, a night vision camera can be used for nocturnal inspections, e.g., an infrared camera with higher contrast than its passive counterpart.

{\small
\bibliographystyle{ieee_fullname}

}

\end{document}